\documentclass[english]{AISB2008}
\usepackage[T1]{fontenc}
\usepackage[latin9]{inputenc}
\usepackage{array}
\usepackage{booktabs}
\usepackage{multirow}
\usepackage{amsmath}
\usepackage{graphicx}

\makeatletter

\providecommand{\tabularnewline}{\\}

\makeatother

\usepackage{babel}
\begin{document}

\title{Towards learning through robotic interaction alone: the joint guided
search task}

\author{Nick DePalma and Cynthia Breazeal\\
Personal Robots Group\\
MIT Media Lab\\
20 Ames Str. Cambridge MA}
\maketitle
\begin{abstract}
This work proposes a biologically inspired approach that focuses on
attention systems that are able to inhibit or constrain what is relevant
at any one moment. We propose a radically new approach to making progress
in human-robot joint attention called \textquotedblleft{}the joint
guided search task\textquotedblright{}. Visual guided search is the
activity of the eye as it saccades from position to position recognizing
objects in each fixation location until the target object is found.
Our research focuses on the exchange of nonverbal behavior toward
changing the fixation location while also performing object recognition.
Our main goal is a very ambitious goal of sharing attention through
probing synthetic foreground maps (i.e. what is being considered by
the robotic agent) and the biological attention system of the human. 
\end{abstract}

\section{Introduction}

As researchers in the field of human-robot interaction begin to make
observations in more longitudinal interactions with participants,
it may need to adapt to new environments, situations, and stimuli
that researchers were not expecting. Success in these domains may
have more to do with the morphology, communicatory competence, and
construction of a synthetic agent than anything else and many of the
underlying problems facing these systems regard adaptation. One of
the most interesting perspectives in embodied adaptation literature
is in the exploration of more situated, sensorimotor behaviors that
are contingent on just the environment itself and not on higher cognitive
processes. Braitenberg \cite{braitenberg1984vehicles} presents a
wonderful introduction to this kind of behavior in the form of thought
experiments. But while these low-level processes may have nice properties
regarding adaptive behavior, they have trouble generalizing and reusing
previous experience. We are inspired by this idea to investigate a
biologically inspired hybrid approach that is constrained by an attention
mechanism to capture a small window of the overall image. By \textit{saccading}
across an image, a fixation window moves and attempts to fixate on
shared positions with the interaction partner. This mechanism allows
its focus of attention to move its window boundaries around objects
and locations for both classification and learning purposes. Our model
investigates a specific class of nonverbal behavior referred to as
\textit{deictic} to direct the fixation point of the attention system. 

Our goal in this work is to build a flexible perception system that
can be used in human-robot interaction domains to extract and learn
about its environment through human interaction alone. Our focus is
on a developmental approach and mechanism called \textit{joint attention}
in which a robot may be directed to attend to something radically
new and still have the capability to refer and learn from this sensor
experience. While our system does also generate goal oriented deictic
action (a critical aspect of \textit{joint} attention), this paper
explores the performance of pixel level referencing vs object level
referencing. In essence, the approach is to extract deictic indices
through the integration of information across multiple modalities
through interactions with the environment and with a social partner.
This triadic relationship between social partner, self, and environment
sets the stage for a more complex cybernetic approach than traditional
robot-environment interactions alone.

This paper documents early ongoing efforts in which we attempt to
apply hand tuned models to correctly predict what other agents are
paying attention to through pure images and gesture alone.

\section{Related Approaches and Positions}

For a robot to share attention with a human participant and vice versa,
it will need to take actions in the world to affect its partners visual
system. Additionally, the robot will need a sensory system that can
handle the actions that are directed at the robot toward predicting
the objects or pixels it is to be directed towards. We draw inspiration
from a number of sources when researching this seemingly simple question,
from epigenetic robotics, human-robot interaction, state of the art
robotic attention systems and basic psychology.

\subsection{Attention and Joint Attention in Developmental Robotics }

\begin{figure}[h]
\begin{centering}
\includegraphics[width=0.2\columnwidth]{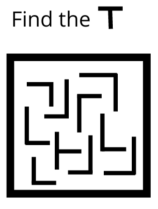}
\par\end{centering}

\caption{Guided search example adapted from Wolfe \cite{wolfe1994guided}.
Guided search is the phenomena of moving the eye around an image to
locate an object. This differs from computational convolution in that
the process may not cover the entire image. }
\label{fig:gs}
\end{figure}

Attention systems can be roughly characterized as bottom-up or top
down. Bottom-up approaches focus primarily on saliency while top-down
attention systems primarily focus on guided search. \textit{Saliency}
can be thought of as intrinsic value of a specific pixel to direct
the fixation location of the agent. \textit{Bottom-up} approaches
focus primarily on understanding how saliency is adapted based on
task and needs of the agent itself. \textit{Top-down} approaches focus
on localizing an object within the image through a particular biased
search mechanism (see Figure \ref{fig:gs}). This is meant to relate
in some way to eye fixation behavior in the computer vision literature.

Since our definition of attention spans both bottom-up and top-down
approaches, we will touch on both approaches to discuss relevant work
in this area. One system that unified saliency maps as a means to
learn new representations between the robot and the environment is
Frintrop's VOCUS embodied attention system \cite{frintrop2006vocus}
which learned about objects via saliency maps and a curiosity system
that was driven to find new and novel objects in its world. VOCUS
was focused primarily on object-environment relations and was not
biased to learn from other agents in its environment. A coverage of
computational attention would not be complete without the decades
long research of Tsotsos\textquoteright{} \cite{tsotsos2011computational}
who presents one theory of computational visual attention based primarily
on Gelade \& Triesman's attention model \cite{treisman1980feature}
that can compute saliency values for arbitrary images. This work is
focused primarily on the computational mechanisms surrounding attention
itself and does not account for social factors. None of these algorithms
focus on robotic joint attention in which the robot plays an active
role in sharing attention with a human participant. 

Scasselatti performed some of the first joint attention work in robotics
\cite{scassellati2001foundations}. The emphasis of this work was
on building a system for a humanoid robot that incorporated a number
of elements of social interaction including a theory of mind, a gaze
following system, and an ecological self. The joint attention system
in this cohesive system followed gaze and pointing gestures toward
a target location but was unable to recognize the object under its
fixation point. Following this work, effort began on learning to follow
gaze from a developmental perspective. Nagai et. al., Doniec et. al.
and Triesch et. al. \cite{nagai2003constructive,triesch2006gaze,doniec2006active}
are all directed at learning how to map referential gesture or gaze
to objects in the world, or in other words, \emph{learning to follow}
a referential gesture toward objects that it already has a model of.
Follow up inquiries about whether or not the robotic visual system
correctly predicted what the human was directing it towards were not
made. Our work attempts to extend previous work by taking a dynamical
systems approach to joint attention that dynamically exchanges gesture
toward sharing attention. We measure the success of our system by
measuring the error of the reported \textit{attentional foreground}
(a mapping of what is inhibited and what is not) and the predicted
attentional foreground of what is being shared.

\subsection{Shared attention through deictics}

Referential gesture is sometimes referred to as \textit{deictic}.
Referential gesture and deictic use in human robot interaction has
been studied in various capacities. \cite{brooks2006working} presents
a model of multimodal deictic generation in communication that leverages
grammar models to generate deictic gesture. \cite{sauppe2014robot}
presents work on specifying a number of categories of deictic use
in reference which include pointing, presenting, touching, exhibiting,
grouping, and sweeping. Other effects such as synchrony \cite{rolf2009attention},
and motionese \cite{nagai2009computational} may also contribute to
the intentional capitalization of innate biases that direct the focus
of attention of the robotic agent. Though joint attention has been
studied in various capacities under different definitions, \cite{kaplan2006challenges}
convincingly argues that the most elusive joint attention phenomena
is the intentional, goal-oriented process and that bottom-up models
where innate attentional biases serendipitously grab the attention
of the interaction group should be considered unintentional shared
attention. Our work focuses primarily on foreground-as-goal and utilizes
deictics as a communicatory action to synchronize foreground.

\section{Joint Guided Search: Task and Implementation}

\begin{figure}[h]
\begin{centering}
\includegraphics[width=0.45\columnwidth]{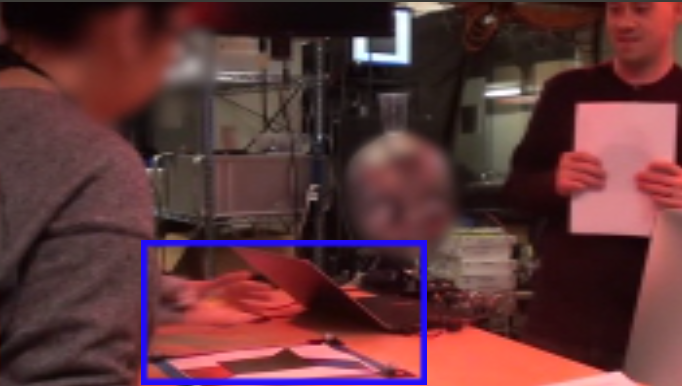}\quad{}\includegraphics[width=0.45\columnwidth]{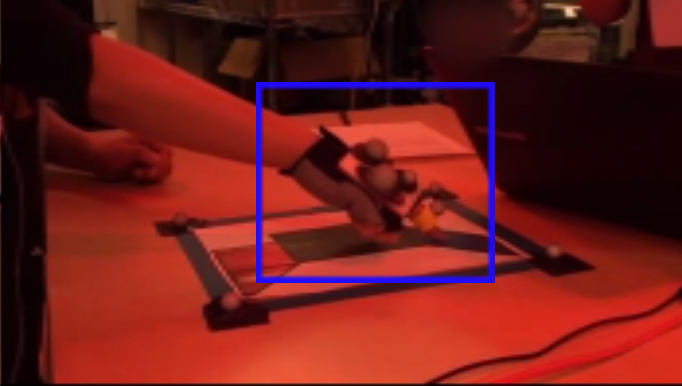}
\par\end{centering}

\caption{Observed behaviors when attempting to direct the attention of another
human participant. Left: a participant using bounded hand gesture
to refer to the space between the palms (highlighted in blue), Right:
precise pointing meant to highlight one particular region that must
be interpreted to be one piece of the tangram figure (highlighted
in blue).}
\label{fig:precisesweep}
\end{figure}

\emph{Joint Guided Search }is the collaborative process by which agents
exchange gesture (in which detection and interpretation of gesture
is the first process which leads to following behavior and prediction
of another's attentional state which we call the \textit{foreground}.
The foreground is a measurement space which may highlight object silhouettes
or highlight the underlying pixel saliency itself. An integrated approach
to improving guided search for agents will require internal robotic
processes to handle mapping symbol or gesture to environment, the
management of appropriate and communicative deictic gesture, and the
interpretation of deictic gesture as directed towards something.

\subsection*{Requirements of a Joint Guided Search Task\label{sub:Requirements-of-a}}

Collaborative joint attention requires that the guided search task
incorporate both predicting the participants foreground and the ability
to take goal-oriented action toward changing the state of attention
of the participant. Foreground is defined as binary maps that represent
the thresholded salience of the scene. This prediction allows the
robot to define a shared attentional space on which both the human
partner and the robot may learn from. Learning through attention mechanisms
is one of the key mechanisms in which learning progresses in biological
agents, but as roboticists, we don't have the technology to support
inquiries into learning from joint visual attention. Our work attempts
to move the state of the art towards these types of inquiries.

A system that can support the demands of joint visual search will
require advances in computer vision and interaction design. Because
this task is behavioral in nature, the internal saccade behavior must
be exposed to the user so that the user may direct the robot to a
more profitable observation positions.

To measure success, we use a normalized mean squared error metric
proposed in DePalma et. al. \cite{depalma2015sensorimotor}. We compare
the predicted foreground map from an image and a deictic action alone.
Section \ref{sec:pilot} describes a pilot study in which a human-human
dyad exchange gesture toward sharing a piece, a part, or the entire
tangram figure. Using the deictic actions collected from this study,
we estimate the foreground from image and action position alone and
compare it to the reported foreground from the observer in the dyad.
We compare the predicted foreground $p^{p}$ with reported foreground
$p^{r}$ using a normalized mean squared error over the image width
$w$ and height $h$:

\begin{align*}
NMSE=\frac{1}{w*h}\sqrt{\sum_{i,j}^{i=1..w,j=1..h}(p_{i,j}^{r}-p_{i,j}^{p})^{2}}
\end{align*}

\subsection{Computational Model of Task Driven Joint Attention}

\begin{figure*}
\begin{centering}
\includegraphics[width=0.7\paperwidth]{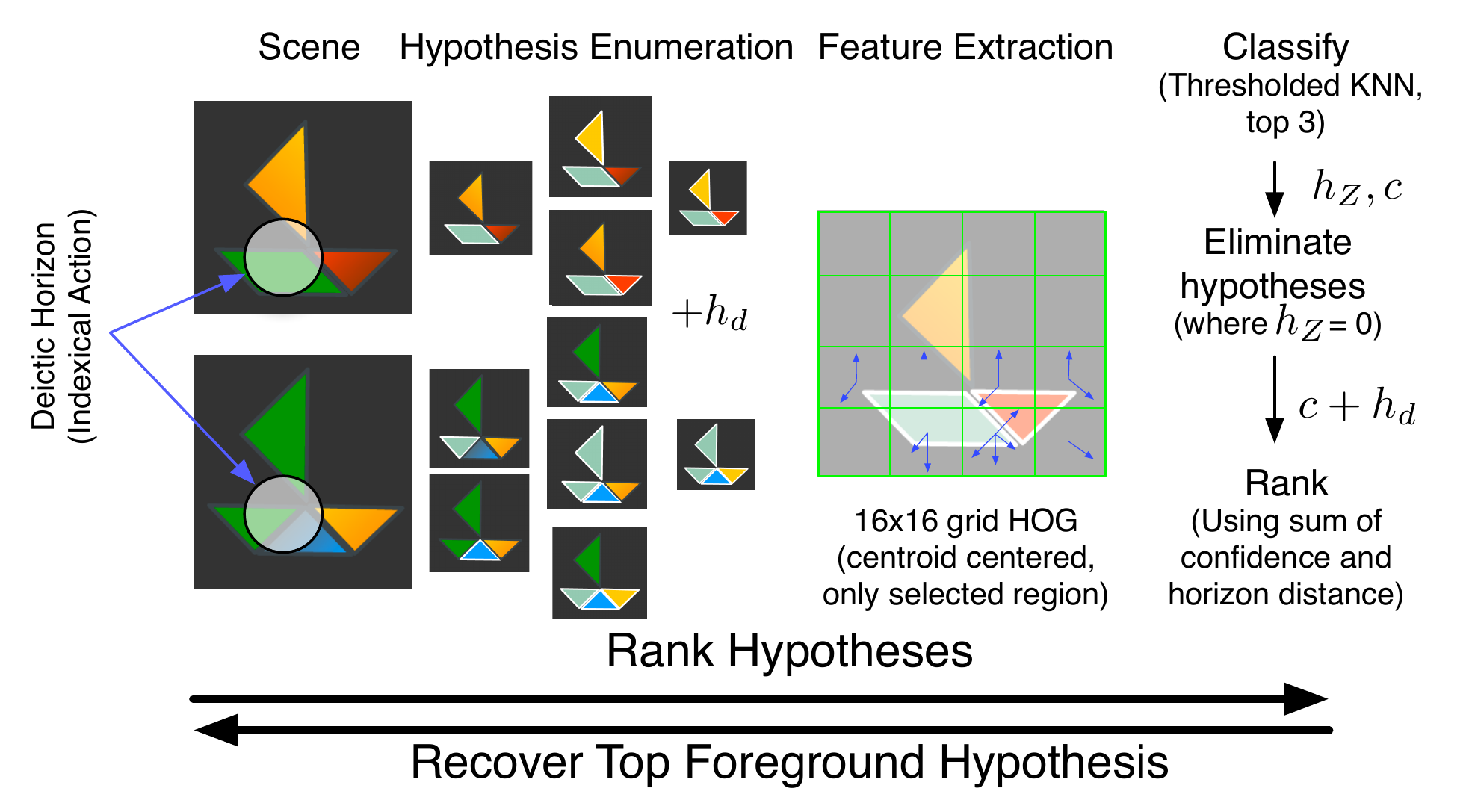}
\par\end{centering}

\caption{Top-down point-ray representation to foreground selection. Object
detection ranking is }

\label{fig:model}
\end{figure*}

The top-down prediction from deictic gesture is computed using a novel
object recognition from fixation point algorithm. The pipeline is
shown in Figure \ref{fig:model}. First, referential action is specified
as $a_{h}(\bar{x},\bar{r},\theta)$, having a point estimate in space
$\bar{x}$, a vector direction $\bar{r}$, and a range (angle $\theta$)
of affected foreground. With known objects $Z=\langle z_{1},z_{2},...,z_{n}\rangle$,
the top down system can classify a current foreground hypothesis as
a known part or object. Note that in this model, $z_{i}$ and $z_{j}$
can (where $0<i,j<n$) have the same label meaning that different
foregrounds can have the same label. 

First, the function projects a cone onto the scene. This cone represents
a horizon boundary in which to enumerate the object hypotheses during
the search for known objects (see Figure \ref{fig:model}). A number
of foregrounds are selected by enumerating all combinations of tangram
pieces whose center points $c_{Z}$ are within the ellipse whose center
is at $c_{a}$. For each foreground hypothesis, the foregrounds are
filtered where $h_{Z}=1$ for the label $Z$. When all of the possible
labels are classified in the given reference region, ranking then
occurs using simple inverse distance $d_{i}(c_{Z},c_{a})=\frac{1}{c+\ell_{2}(c_{Z},c_{a})}$
where $\ell_{2}$ represents the L2-norm. $c_{Z}$ is calculated by
taking the centroid of the foreground in which $h_{Z}=1$. To train
the $h_{Z}$ classifiers, we used HOG features \cite{dalal2005histograms}
from the rastered images of the tangrams. 

The resulting foregrounds that underly the symbol form the predicted
set of potential reference foregrounds, $\langle F_{Z}\rangle$. The
top ranking prediction (with smallest error), $F_{Z}=argmin_{d}d(c_{Z},c_{a})$,
is used as the top-down contribution.

\section{Data Collection, Pilot Task}

\label{sec:pilot}

\begin{figure}[th]
\begin{centering}
\begin{tabular}{c}
\includegraphics[width=1\columnwidth]{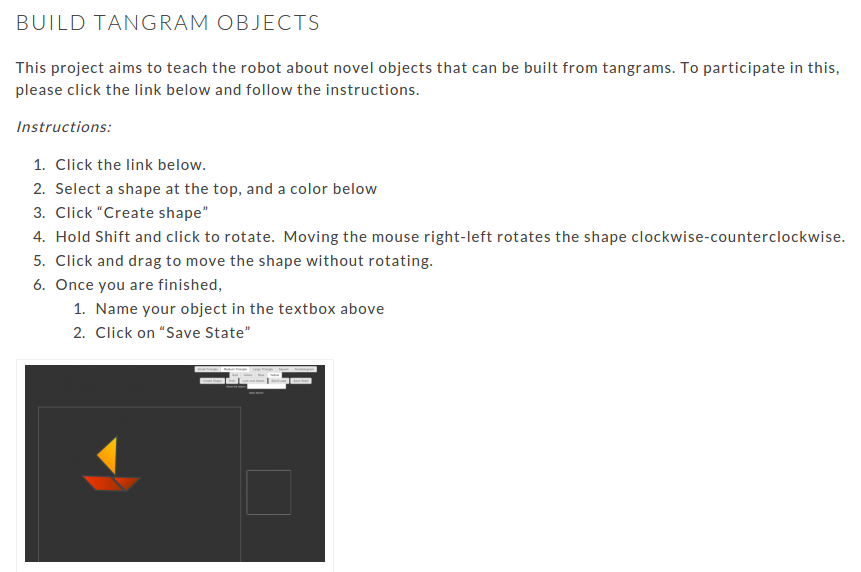}\tabularnewline
\includegraphics[width=0.3\columnwidth]{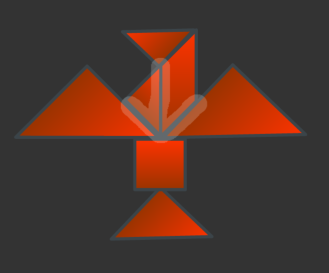}\quad{}\includegraphics[width=0.3\columnwidth]{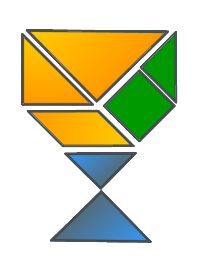}\quad{}\includegraphics[width=0.3\columnwidth]{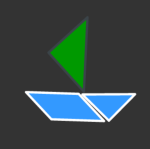}\tabularnewline
\end{tabular}
\par\end{centering}

\caption{Top: Crowdsourcing based online interface. Bottom Left: low-level
tangram foreground collected online (pixel based). Bottom Middle:
Tangram figure taken from our dataset collected online. Bottom Right:
high-level tangram foreground collected online (object based)}

\label{lab:task}
\end{figure}

To understand when referential gesture refers to a part of the scene
that is unknown or whether the reference should map to something previously
known, we devised a tangram task in which the goal of the reference
could be very low level maps (Figure \ref{lab:task}) or higher level
structures (Figure \ref{lab:task}). Our goal is to minimize the measure
of error between the predicted foreground and the goal foreground
of each referential action exchanged between a human dyad sharing
a scene. A scene that was collected online is presented to the dyad
and roles are given to each participant. The setup (pictured in Figure
\ref{fig:precisesweep}) includes a shared scene composed of tangrams
. One participant of the dyad is assigned the role of \textit{showing}
the participant what foreground they must enter in their touchscreen
without any verbal communication. The \textit{observer }then observes
the gestures and then returns to their touchscreen to enter the foreground
into the image. 

We first began by collecting a wide range of tangrams and tangram
goals online (Table \ref{lab:task}). The basic task of collecting
our dataset was to begin by asking users to provide tangram figures
of their choice through online play with the system. Finally, they
are allowed to select the parts by clicking on the pieces and labeling
them (e.g. they can select arms of a man, heads of bird, etc). Finally,
a secondary task was provided to random users on the internet in which
we asked them to highlight the low-level regions of the figure that
they found most interesting and those foregrounds are used as low-level
goals. A total of 5 dyads were collected across 30 scenes, collecting
a total of 150 total scenes in which interaction was observed.

\section*{Results}

\label{sec:Results}

We first separate the goals into two datasets: low level, unknown
goals (UG dataset) and high-level, known goals (KG dataset). We then
analyzed and collected the set of actions from the human dyad dataset
in which we found that within a single dyad session, either one single
gesture was exchanged (SG) or multiple referential gestures are exchanged
(MG). For each of those groups, we isolated the datasets into foreground
goals in which the system had an object that it could predict and
those in which the foreground was pixel based to compare the advantages
and disadvantages of each situation that a robot may encounter.

\begin{figure}[h]
\begin{centering}
\includegraphics[width=0.9\columnwidth]{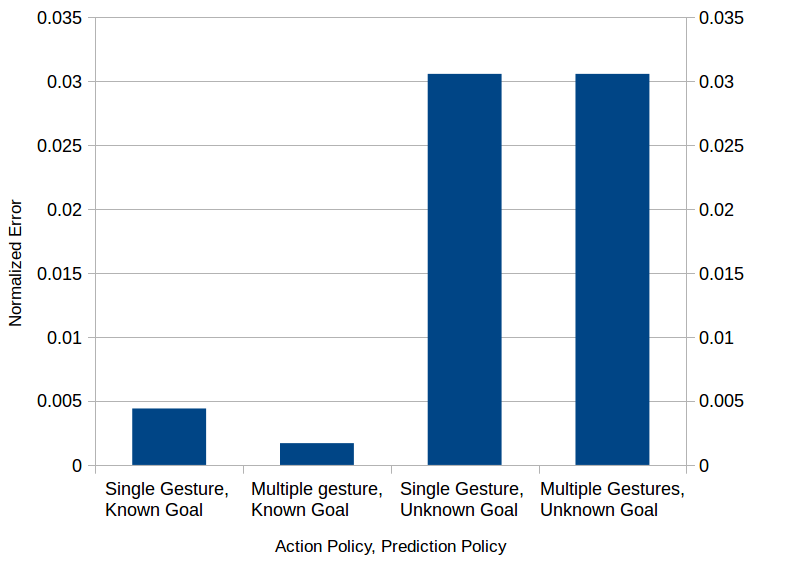}
\par\end{centering}

\caption{Foreground prediction performance of our guided search mechanism compared
against multiple observed action strategies.}

\label{fig:topdownres}
\end{figure}

\begin{table}
\begin{centering}
\begin{tabular}{|c|c|c|c|c|}
\hline 
\multirow{2}{*}{} & \multicolumn{3}{c|}{Reported significance} & \multirow{2}{*}{NMSE}\tabularnewline
\cline{2-4} 
 & MGKG & SGUG & MGUG & \tabularnewline
\hline 
\hline 
SGKG & p < 0.06 & p < 0.001 & p < 0.001 & 0.004\tabularnewline
\hline 
MGKG &  & p < 0.001 & p < 0.001 & 0.001\tabularnewline
\cline{1-1} \cline{3-5} 
SGUG & \multicolumn{2}{c|}{} & p > 0.9 & 0.03\tabularnewline
\cline{1-1} \cline{4-5} 
MGUG & \multicolumn{3}{c|}{} & 0.03\tabularnewline
\cline{1-1} \cline{5-5} 
\end{tabular}
\par\end{centering}

\caption{Reported significance and error of a single foreground proposer method
against known goals and unknown goals. Significance values are reported
using a Student's $t$-test and the average normalized mean squared
error is reported for each dataset on the far right.}
\end{table}

For this paper, we report the results regarding the robot's ability
to predict the foreground of the human participant. The goal image
foreground and the predicted image foreground were aligned and the
normalized mean squared error (see Section \ref{sub:Requirements-of-a})
was reported on the y-axis. Figure \ref{fig:topdownres} shows a clear
performance difference between deictic gesture to scene prediction
in which the goal is already known but had to be registered on the
scene and the goals which have no top-down object based representation
that could be used to improve the performance of the foreground prediction.
Reported $p$-values using a Student's $t$-test between each group
show that the top-down visual proposer is not enough to predict foreground
and that an agent will need to balance foreground prediction between
well known object predictors and bottom-up pixel highlighting. It
is clear that we will need some type of insight into foreground prediction
that allows the agent to saccade to unknown stimuli so that the agent
may build new object representations on the fly. 

We are encouraged by these results and are moving forward to extend
the system to understand when new stimuli are encountered, how to
best make a prediction on the foreground and then to make clarifying
gestures with the robot that will allow us to improve foreground prediction
over the course of an interaction. We are also extending this system
to build visual representations dynamically through the interaction.
In the long run, we hope to understand how behavior influences the
representations that emerge.

\section*{Future Work}

Our future plan with this work is to report on the other proposers
influence on the foreground prediction as well as whether or not the
robot's gesture can allow the human to predict what it is the robot
desires the human to attend to. We are also interested in understanding
how social behavior can influence the learned representations and
compare them against representations that were learned through robot-environment
actions alone. There is much work in looking at robot-environment
interactions and their influence on the representations. VOCUS \cite{frintrop2006vocus}
has reported the most complete results from a system like this but
again, they are not focused on the interaction domain. Additionally,
recent work in learning through attention in neural networks have
shown very positive results but are not biased by social factors \cite{tang2014learning}.
We are also interested in extending this work to real world domains
once these learning systems are able to operate in more interactive
domains. 

\bibliographystyle{IEEEtran}
\bibliography{attention}

\begin{thebibliography}{10}
\providecommand{\url}[1]{#1}
\csname url@samestyle\endcsname
\providecommand{\newblock}{\relax}
\providecommand{\bibinfo}[2]{#2}
\providecommand{\BIBentrySTDinterwordspacing}{\spaceskip=0pt\relax}
\providecommand{\BIBentryALTinterwordstretchfactor}{4}
\providecommand{\BIBentryALTinterwordspacing}{\spaceskip=\fontdimen2\font plus
\BIBentryALTinterwordstretchfactor\fontdimen3\font minus
  \fontdimen4\font\relax}
\providecommand{\BIBforeignlanguage}[2]{{%
\expandafter\ifx\csname l@#1\endcsname\relax
\typeout{** WARNING: IEEEtran.bst: No hyphenation pattern has been}%
\typeout{** loaded for the language `#1'. Using the pattern for}%
\typeout{** the default language instead.}%
\else
\language=\csname l@#1\endcsname
\fi
#2}}
\providecommand{\BIBdecl}{\relax}
\BIBdecl

\bibitem{braitenberg1984vehicles}
V.~Braitenberg, \emph{Vehicles: Experiments in synthetic psychology}.\hskip 1em
  plus 0.5em minus 0.4em\relax MIT press, 1984.

\bibitem{wolfe1994guided}
J.~M. Wolfe, ``Guided search 2.0 a revised model of visual search,''
  \emph{Psychonomic bulletin \& review}, vol.~1, no.~2, pp. 202--238, 1994.

\bibitem{frintrop2006vocus}
S.~Frintrop, \emph{VOCUS: A visual attention system for object detection and
  goal-directed search}.\hskip 1em plus 0.5em minus 0.4em\relax Springer, 2006,
  vol. 3899.

\bibitem{tsotsos2011computational}
J.~K. Tsotsos, \emph{A computational perspective on visual attention}.\hskip
  1em plus 0.5em minus 0.4em\relax MIT Press, 2011.

\bibitem{treisman1980feature}
A.~M. Treisman and G.~Gelade, ``A feature-integration theory of attention,''
  \emph{Cognitive psychology}, vol.~12, no.~1, pp. 97--136, 1980.

\bibitem{scassellati2001foundations}
B.~Scassellati, ``Foundations for a theory of mind for a humanoid robot,''
  Ph.D. dissertation, Massachusetts Institute of Technology, 2001.

\bibitem{nagai2003constructive}
Y.~Nagai, K.~Hosoda, A.~Morita, and M.~Asada, ``A constructive model for the
  development of joint attention,'' \emph{Connection Science}, vol.~15, no.~4,
  pp. 211--229, 2003.

\bibitem{triesch2006gaze}
J.~Triesch, C.~Teuscher, G.~O. De{\'a}k, and E.~Carlson, ``Gaze following: why
  (not) learn it?'' \emph{Developmental science}, vol.~9, no.~2, pp. 125--147,
  2006.

\bibitem{doniec2006active}
M.~W. Doniec, G.~Sun, and B.~Scassellati, ``Active learning of joint
  attention,'' in \emph{Humanoid Robots, 2006 6th IEEE-RAS International
  Conference on}.\hskip 1em plus 0.5em minus 0.4em\relax IEEE, 2006, pp.
  34--39.

\bibitem{brooks2006working}
A.~G. Brooks and C.~Breazeal, ``Working with robots and objects: Revisiting
  deictic reference for achieving spatial common ground,'' in \emph{Proceedings
  of the 1st ACM SIGCHI/SIGART Conference on Human-robot Interaction}.\hskip
  1em plus 0.5em minus 0.4em\relax ACM, 2006, pp. 297--304.

\bibitem{sauppe2014robot}
A.~Saupp{\'e} and B.~Mutlu, ``Robot deictics: How gesture and context shape
  referential communication,'' in \emph{Proceedings of the 2014 ACM/IEEE
  international conference on Human-robot interaction}.\hskip 1em plus 0.5em
  minus 0.4em\relax ACM, 2014, pp. 342--349.

\bibitem{rolf2009attention}
M.~Rolf, M.~Hanheide, and K.~J. Rohlfing, ``Attention via synchrony: Making use
  of multimodal cues in social learning,'' \emph{Autonomous Mental Development,
  IEEE Transactions on}, vol.~1, no.~1, pp. 55--67, 2009.

\bibitem{nagai2009computational}
Y.~Nagai and K.~J. Rohlfing, ``Computational analysis of motionese toward
  scaffolding robot action learning,'' \emph{Autonomous Mental Development,
  IEEE Transactions on}, vol.~1, no.~1, pp. 44--54, 2009.

\bibitem{kaplan2006challenges}
F.~Kaplan and V.~V. Hafner, ``The challenges of joint attention,''
  \emph{Interaction Studies}, vol.~7, no.~2, pp. 135--169, 2006.

\bibitem{depalma2015sensorimotor}
N.~Depalma and C.~Breazeal, ``Sensorimotor account of attention sharing in hri:
  Survey and metric,'' \emph{2nd Annual Symposium on Artificial Intelligence
  and Human-Robot Interaction}, 2015.

\bibitem{dalal2005histograms}
N.~Dalal and B.~Triggs, ``Histograms of oriented gradients for human
  detection,'' in \emph{Computer Vision and Pattern Recognition, 2005. CVPR
  2005. IEEE Computer Society Conference on}, vol.~1.\hskip 1em plus 0.5em
  minus 0.4em\relax IEEE, 2005, pp. 886--893.

\bibitem{tang2014learning}
Y.~Tang, N.~Srivastava, and R.~R. Salakhutdinov, ``Learning generative models
  with visual attention,'' in \emph{Advances in Neural Information Processing
  Systems}, 2014, pp. 1808--1816.

\end{thebibliography}

\end{document}